\documentclass[10pt,twocolumn,letterpaper]{article}

\usepackage{iccv}
\usepackage{times}
\usepackage{epsfig}
\usepackage{graphicx}
\usepackage{amsmath}
\usepackage{amssymb}

\usepackage[breaklinks=true,bookmarks=false]{hyperref}

\iccvfinalcopy 

\def\iccvPaperID{****} 

\begin{document}

\title{Towards Purely Unsupervised Disentanglement of Appearance and Shape for Person Images Generation}

\author{Hongtao Yang\\
Australian National University\\
{\tt\small hongtao.yang@anu.edu.au}
\and
Tong Zhang\\
Australian National University\\
{\tt\small tong.zhang@anu.edu.au}
\and
Wenbing Huang\\
Tsinghua University\\
{\tt\small hwenbing@126.com}
\and
Xuming He\\
ShanghaiTech University\\
{\tt\small hexm@shanghaitech.edu.cn}
\and
Fatih Porikli\\
Australian National University\\
{\tt\small fatih.porikli@anu.edu.au}
}

\def\eg{\emph{e.g}\onedot} \def\Eg{\emph{E.g}\onedot}
\def\ie{\emph{i.e}\onedot} \def\Ie{\emph{I.e}\onedot}
\def\Mat#1{{\boldsymbol{#1}}}
\maketitle
\maketitle

\begin{abstract}
There have been a fairly of research interests in exploring the disentanglement of appearance and shape from human images. Most existing endeavours pursuit this goal by either using training images with annotations or regulating the training process with external clues such as human skeleton, body segmentation or cloth patches etc.
In this paper, we aim to address this challenge in a more unsupervised manner---we do not require any annotation nor any external task-specific clues. 
To this end, we formulate an encoder-decoder-like network to extract both the shape and appearance features from input images at the same time, and train the parameters by 
three losses: feature adversarial loss, color consistency loss and reconstruction loss. The feature adversarial loss mainly impose little to none mutual information between the extracted shape and appearance features, 
while the color consistency loss is to encourage the invariance of person appearance conditioned on different shapes. More importantly, our unsupervised\footnote{Unsupervised learning has many interpretations in different tasks. To be clear, in this paper, we refer unsupervised learning as learning without task-specific human annotations, pairs or any form of weak supervision.} framework utilizes learned shape features as masks which are applied to the input itself in order to obtain clean appearance features. Without using fixed input human skeleton, our network better preserves the conditional human posture while requiring less supervision. 
Experimental results on DeepFashion and Market1501 demonstrate that the proposed method achieves clean disentanglement and is able to synthesis novel images of comparable quality with state-of-the-art weakly-supervised or even supervised methods.
\end{abstract}

\begin{figure}[t]
	\begin{center}
		\includegraphics[width=1.0\linewidth]{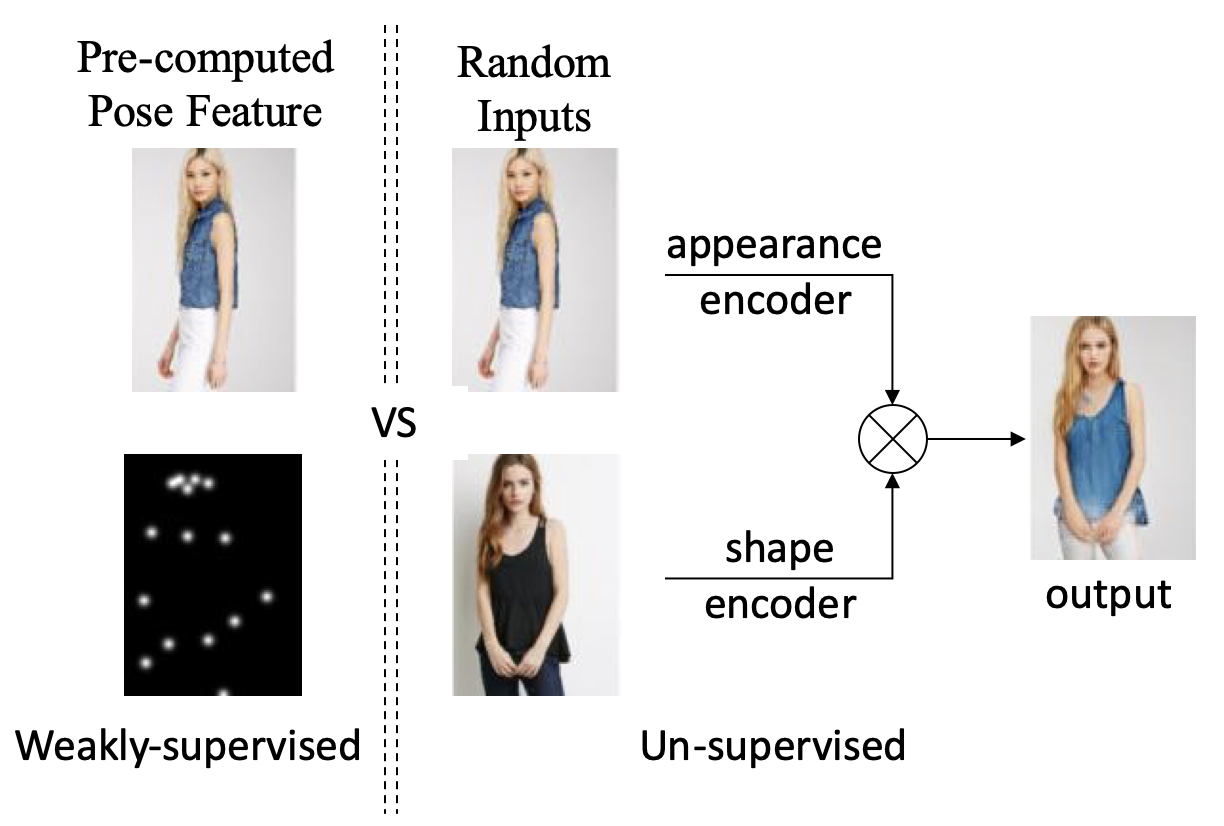}
	\end{center}
	\caption{\textbf{Left}: The problem setup of existing methods. The challenge is to learn disentangled appearance features given an image and its corresponding shape representation. We call this \emph{weakly-supervised disentanglement}. \textbf{Right}: The problem setup of our approach. The challenge is to learn disentangled appearance and shape representations given only an image, without any task-specific cues. We call this \emph{unsupervised disentanglement}. The output in this figure is a novel image generated by our framework.}
	\label{fig.problem_setup}
	\vspace{-2mm}
\end{figure}

\vspace{-2mm}
\section{Introduction}
How to automatically, without any supervision, learn and decompose the characteristics of an object in the world is an ultimate goal in computer vision. Conversely, learn to synthesize components from different objects and generate novel images is also very challenging. This problem has attracted an extraordinary amount of interests recently due to rapid progress in deep generative models~\cite{kingma2013auto,oord2016pixel,goodfellow2014generative}.

While a variety of generative frameworks have been proposed, the generative adversarial network (GAN)~\cite{goodfellow2014generative} and its variants~\cite{arjovsky2017wasserstein,radford2015unsupervised,gulrajani2017improved} have arguably become most prevalent due to high-quality images they can produce.  Through different learning strategies on such models~\cite{mirza2014conditional,isola2017image,ma2017pose,zhang2017stackgan++}, the well-trained latent representation of images are able to synthesize novel images with desirable properties according to the external inputs on latent space. Unfortunately, these methods need laborious labeling, pairing or other supervision signals to enforce the network to memorize the mapping. More importantly, their performance are far from satisfactory in this task. At the same time, unsupervised disentangling methods~\cite{chen2016infogan,higgins2017beta} currently are only able to be applied in very naive datasets. It is impractical to apply unsupervised learning do disentangle all the characteristics.

In this paper, in order to generate realistic and novel human images unsupervisedly, we narrow down components to the most prominent two properties of human: \emph{appearance} and \emph{shape}. Learning a latent representation that properly disentangles those two factors from images, however, still remains a challenging task. Previous attempts~\cite{ma2017pose} \cite{isola2017image,villegas2017learning,fan2018controllable,siarohin2018deformable,dong2018soft,han2018viton,wang2018toward} to this task have always resorted to paired data (\emph{e.g.} two images with the same appearance but different poses) in order to obtain ground-truth supervision on what generated images should look like. While in limited scenarios where the paired information is easy to acquire \cite{villegas2017learning}, obtaining paired data is often expensive, time-consuming and may require extra complex operations \cite{chan2018everybody}. 

Although a few attempts~\cite{esser2018variational,ma2018disentangled,raj2018swapnet,pumarola2018unsupervised} have managed to address this problem using only unpaired data by leveraging the recent success of unpaired image to image translation algorithms~\cite{zhu2017unpaired,choi2017stargan,liu2017unsupervised,bousmalis2017unsupervised,liu2016coupled,taigman2016unsupervised,kim2017learning,pumarola2018ganimation,yi2017dualgan,lee2018diverse}, they always require explicit attribute-specific cue as additional inputs, such as human skeleton~\cite{esser2018variational,pumarola2018unsupervised}, joint heatmaps~\cite{ma2018disentangled}, body segmentation~\cite{raj2018swapnet,mo2018instagan}, cloth patches~\cite{han2018viton,wang2018toward} or a combination of several cues \cite{dong2018soft}, in order to individually control the two properties of the generated images and looking realistic at the same time. While these explicit attribute-specific cues can provide valuable supervision information, they are fixed and manually defined, thus exposing two limitations:
\begin{enumerate}
    \item Complex pre-processing and heavy engineering tricks are often employed in order to utilize the attribute-specific cues. Examples include: convert joint coordinates to heatmaps using 2D Gaussian noise~\cite{villegas2017learning}; patching white circles of specific radius around each joints locations~\cite{ma2017pose,ma2018disentangled}; carefully drawing stickman~\cite{esser2018variational,pumarola2018unsupervised}; and manually crop or warp appearances patches around every body parts~\cite{ma2018disentangled,esser2018variational}. Not only do those pre-processing severely limit the flexibility of their methods, they also impose strong prior feature representations rather than learning from data. In other words, using explicit attribute-specific cues limits the network's ability of leaning plausible disentangled representations in shape and appearance space.
    \item The predefined attribute-specific cues inevitably carry undesired bias into the system, which will hinder the quality of the generated images. For example, artificially defined human poses such as skeleton or joint coordinates cannot comprehensively describe the shape of a person. Such shape information lack important details about \emph{e.g.} thickness of limbs, hair styles etc, resulting in stiff and unnatural looking generated persons~\cite{esser2018variational}. Another example is that body segmentation obtained with human parsing~\cite{gong2017look,gong2018instance} often focus on types of apparels rather than the underlying human structure, restricting the use case to only a few specific types of apparel~\cite{mo2018instagan}. Therefore, usage of the attribute-specific cues absent of further refinements may result in ill-suited representations for image translation task.
\end{enumerate}
 
We argue that a more general learning strategy is to extract more plausible data-driven disentangled representations for appearance and shape respectively through unsupervised learning. Figure \ref{fig.problem_setup} illustrates our problem setup and motivation. We aim to automatically infer better appearance and shape features from two randomly chosen images, and generate novel image accordingly. To this end, we propose an end-to-end unsupervised network integrating prior knowledge of human pose detection, to automatically extract disentangled representations and generate novel images. The human pose prior is adaptively refined with our training objectives, producing a more comprehensive shape descriptor. We argue that the learned shape and appearance descriptors, compared to explicit attribute cues, produce cleaner feature representations and are better suited for the task of human shape and appearance transfer. Also, not requiring any extra information input makes our network more flexible and widely applicable.

Specifically, our proposed method explores cycle consistency and adversarial training to simultaneously encourage independence and high expressive power of the learned features. The network consists of three components: A shape encoder that generates a set of masks corresponds to different body parts; An appearance encoder that capture appearance characteristics based on the shape masks; And finally a image decoder that takes the concatenated input of appearance features and shape masks to generate realistic images with desired attributes.

We demonstrate the capability of our proposed method on two different datasets: DeepFashion~\cite{liuLQWTcvpr16DeepFashion} and  Market1501~\cite{zheng2015scalable}. We obtain convincing results on both datasets, with favorable comparisons to the state-of-the-art approaches. Our main contributions are summarized as follows:
\begin{itemize}
\vspace{-2mm}
    \item We propose an end-to-end network to learn disentangled representations of appearance and shape in an unsupervised manner.
    \item Our unsupervised framework for conditional appearance and shape transfer eliminate the need for explicit attribute cues inputs.
    \item Our extensive experiments demonstrate that our unsupervised framework outperform state-of-the-arts algorithms quantitatively and visually, including some supervised and weakly-supervised methods.
   
\end{itemize}
\section{Related Work}
 Promising results have been accomplished under supervised setting when input-output images are paired exactly. The most successful supervised method is the Pix2pix framework \cite{isola2017image,wang2017high}, which applies a conditional discriminator \cite{mirza2014conditional} on image patches to generate high quality images. However, the problem becomes more challenging with no available paired data. A major limitation in previous unpaired image translation methods~\cite{zhu2017unpaired,choi2017stargan,liu2017unsupervised,bousmalis2017unsupervised,liu2016coupled,taigman2016unsupervised,kim2017learning,pumarola2018ganimation,yi2017dualgan,lee2018diverse} is that they can not manipulate attributes independently to generate novel images, severely restricting their use case to image with the same spatial structures. When spatial structures differs dramatically between inputs and outputs, as is the case in human image translation, they often produce unpleasant results cluttered with visual artifacts (See Figure~\ref{fig.compare_deepfashion} and \ref{fig.compare_market}). As a consequence, we can not directly apply similar methods on the task of conditional appearance and shape transfer. Instead, we resort to learning disentangled representations to manipulate image attributes and achieve realistic generation results.

\subsection{Unpaired Human Image Generation}
A fair amount of works have been proposed to tackle unpaired appearance and shape transfer~\cite{esser2018variational,ma2018disentangled,raj2018swapnet,pumarola2018unsupervised}. However, they all rely on to some type of additional information such as joint coordinates~\cite{esser2018variational,pumarola2018unsupervised,ma2018disentangled} or stand-alone cloth patches~\cite{han2018viton,wang2018toward}. We argue they only managed to address the task in a weakly-supervised manner. Mo \cite{mo2018instagan} took a different approach by exploring domain-specific information, and propose instance-aware transfer using human parsing masks. However, their model can only be applied on specific types of clothes and needs to be retrained for other apparels, making it unpractical.



\subsection{Disentangled Representation}
Learning disentangled representation has been a pursuit of machine learning researchers. With the recent developments of generative models, many methods are able to discover factors of variation other than those relevant for classification. Mathieu~\cite{mathieu2016disentangling} utilizes a set of labeled observations to discover other factors that are independent to the label through cross-domain image translation and reconstruction. \cite{yim2015rotating,peng2017reconstruction,liu2018multi} use images of different styles to learn a disentangled factor that control the style. Taking advantage of the temporal coherence and the rich pairwise information in videos, \cite{denton2017unsupervised,villegas2017decomposing,baddar2017dynamics,tulyakov2017mocogan} attempts to factor the video into temporal stationary and temporal varying components, or content and motion. The above works all try to acquire other factors of variation with the help of a given factor, such as the styles of the image or the poses of the frames. It is more challenging to acquire disentangled representations in an unsupervised manner. The method by~\cite{hu2017disentangling} manages to automatically discover different factors by mixing and unfolding latent representations. We absorb inspirations from \cite{hu2017disentangling} but focus on mixing the appearance and pose features in our paper.

\begin{figure*}[t]
	\begin{center}
		\includegraphics[width=0.85\linewidth, height=0.42\linewidth]{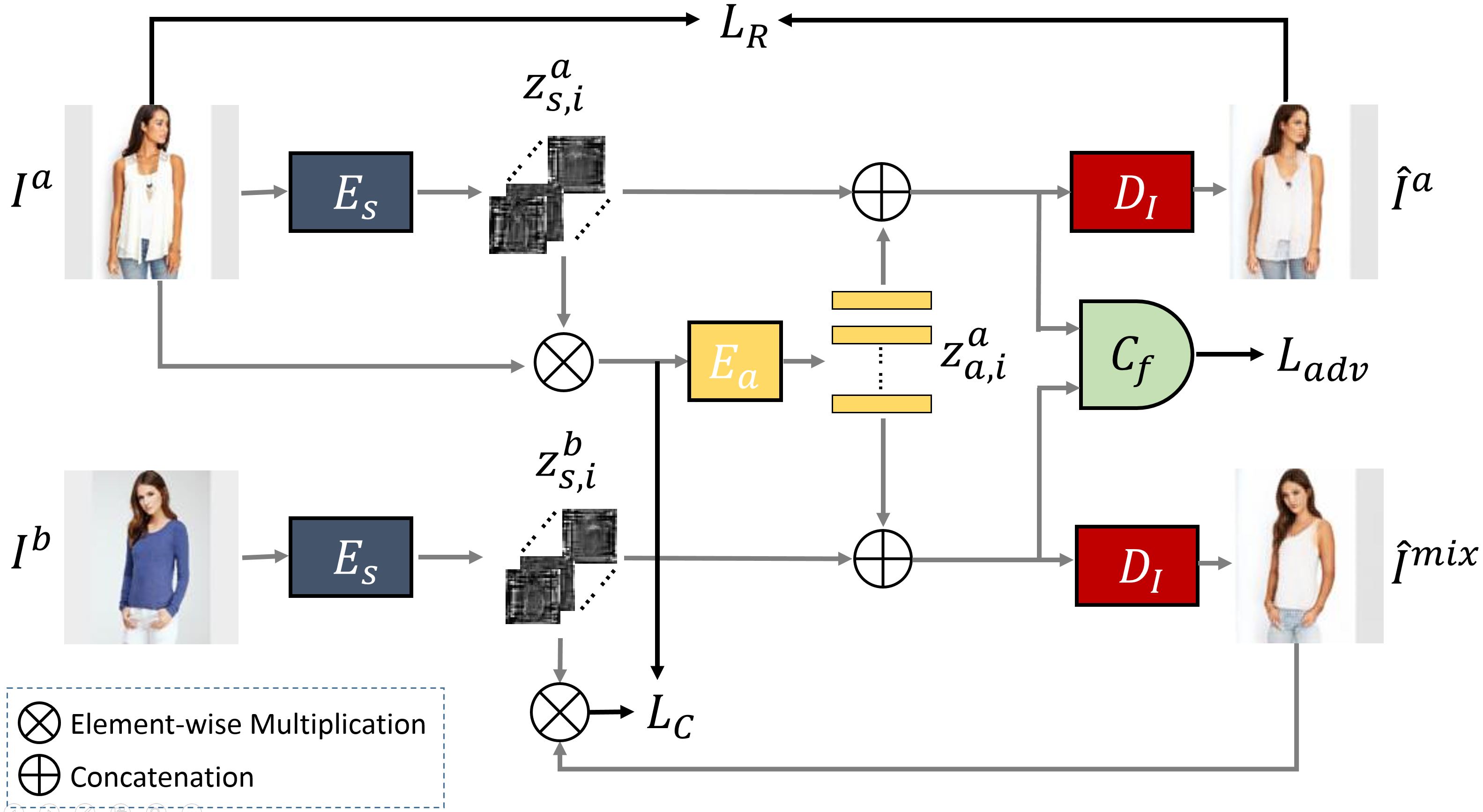}
	\end{center}
	\vspace{-3mm}
	\caption{The overview of the network.}
	\label{fig.network_overview}
	\vspace{-3mm}
\end{figure*}

\vspace{-2mm}
\section{Method}
As illustrated in Figure~\ref{fig.problem_setup}, our goal is to generate novel realistic images based on two input images that respectively supply the desired appearance and shape conditions. We achieve this in an unsupervised manner where neither the ground truths of the output nor attribute-specific cues (\emph{e.g.} human skeletons) are needed. The model we learn is able to automatically disentangle shape from appearance, and thus can generate images under arbitrary combination of them.

Formally, let us denote by $\Mat{I}^a,\Mat{I}^b \in R^{H\times W\times 3}$ the two RGB images of the height $H$ and width $W$. We suppose the shape and appearance representations for image $\Mat{I}^a$ to be  $\Mat{z}_s^a$ and  $\Mat{z}_a^a$, respectively, and define the features $\Mat{z}_s^b$ and $\Mat{z}_a^b$ of $\Mat{I}^b$ similarly. 
Our task is thus to learn an image decoder $D_I(\Mat{z}_a^a, \Mat{z}_s^b)$ that combines the appearance from $\Mat{I}^a$ with the shape of $\Mat{I}^b$ to produce a novel image $\Mat{\hat{I}}^{mix}$ when $a\neq b$.
The core of our method is how to encourage the disentanglement between the shape and appearance features, thus we can expect a realistic looking mixed image if we retain either of them but freely alter the other.
As discussed before, we strive to address the problem with no external supervision, hence how to train our model requires specific network design, which is the topic of this section. 
We first present details of the network, then define the objective functions and explain our training strategy.

\subsection{Architecture Design}\label{sec:network}
The network consists of four modules: 1. a shape encoder $E_s$ that generates a set of shape masks; 2. an appearance encoder $E_a$ that encodes appearance contents attended by the shape masks from $E_s$; 3. an image decoder $D_I$ that generates photo-realistic images based on the concatenation of the shape and appearance features; 4. a feature classifier that learns to classify whether the given appearance and shape features are from the same image or not. The overview of the network is shown in Figure~\ref{fig.network_overview}. We defer more details of the network structure into the supplementary material.

\paragraph{Shape and Appearance Encoders} The shape encoder $E_s$ is a convolutional network that down-samples the input image $\Mat{I}^a$ and emit a set of $m$ masks $\{\Mat{z}_{s,i}^a, \ i=1...m\}$, where the value of each pixel in $\Mat{z}_{s,i}^a$ is within $[0,1]$. These masks serve dual purposes: one to filter local regions of the input images and the other one to provide conditional shape input to the image decoder $D_I$. Particularly for the first purpose, each of the $m$ masks is first resized spatially and broadcasted along channels to match the size of the input image, and then is applied to the input image by element-wise multiplication, thus resulting in $n$ filtered images. All these filtered images are passed through the appearance encoder $E_a$ (also a down-sampling convolution network) to deliver $n$ appearance feature vectors, namely, $\Mat{z}_a^a=\{\Mat{z}_{a,1}^a, \Mat{z}_{a,2}^a ... \Mat{z}_{a,m}^a\}$.
It should be noted that, the fusion of the shape and input images prior to the input of the appearance encoder makes our method clearly distinct from those conventional approaches~\cite{pumarola2018unsupervised,esser2018variational} where the encoding processes of both attributes are independent to each other. We contend that, the spatial-attended fusion is reasonable and efficient, as the shape mask should be naturally combinative to the raw image content and the filtered image could focus more on the meaningful regions of the person. 

\paragraph{Image Decoder} The shape and appearance representations $\Mat{z}_s$ and $\Mat{z}_a$ are concatenated as input to the image decoder $D_I$ for image decoding. As is shown in Figure~\ref{fig.network_overview}, we simultaneously decode reconstruction image $\Mat{\hat{I}}^a = D_I(\Mat{z}_s^a, \Mat{z}_a^a)$ and generate novel image $\Mat{\hat{I}}^{mix} = D_I(\Mat{z}_s^b, \Mat{z}_a^a)$ using their corresponding features. The reconstructed $\Mat{\hat{I}}^a$ is associated with an reconstruction loss to satisfy the cycle consistency, while the novel $\Mat{\hat{I}}^{mix}$ is used to compute the color consistency loss. Details are explained in Section~\ref{sec:model_learning}.

\paragraph{Feature Classifier} 
Another core assumption of our method is that the distributions of the shape masks extracted from different persons should be consistent and identical. This assumption is crucial, as it enables the transferring ability of the pose from one person to another. To do so, we propose a feature classifier $C_f$ which takes as input the concatenation of the shape and appearance representations, and determines if they are from the same person. For instance in Fig~\ref{fig.network_overview}, we have the true pair $[\Mat{z}_s^a, \Mat{z}_a^a]$  and the false one $[\Mat{z}_s^b, \Mat{z}_a^a]$. 
The classifier will output $1$ when $a=b$ and $0$ otherwise.
We then train the classifier and the encoder in an adversarial way to make close the distributions of the shape masks $\Mat{z}_s^a$ and $\Mat{z}_s^b$.

\subsection{Model Training}\label{sec:model_learning}
Motivated by the discussions above, we apply three losses: the \emph{reconstruction loss} that trains the network to generate photo-realistic images; the \emph{feature adversarial loss} that ensures universal representative ability of the shape mask; and the \emph{color consistency loss} that encourages consistent appearance transfer. Below we describe them in details.

\vspace{-4mm}
\paragraph{Reconstruction Loss} Conventional pixel-wise reconstruction loss such as L1-norm enforces exact alignments between two images. This yet is too rigorous for our framework because certain minor granularity is inevitably lost if we attempt to disentangle shape from appearance. Therefore, we adopt the image perceptual loss \cite{chen2017photographic,johnson2016perceptual} instead to constrain the reconstructed image to be perceptually the same as the original. Formally, the reconstruction loss is:
\begin{eqnarray}
\mathcal{L}_R = \sum_k \phi_k ||\Phi_k(\Mat{I}^a) - \Phi_k(D_I(\Mat{z}_s^a, \Mat{z}_a^a))||_1
\vspace{-3mm}
\end{eqnarray}
where $\Phi$ is the VGG \cite{simonyan2014very} network used to compute the perceptual similarities measured at the $k$th layer, and $\phi_k$ controls its weight.

\paragraph{Feature Adversarial Loss} We impose this loss via an adversarial training between the feature classifier $C_f$ and the shape/appearance encoders $E_s$ and $E_a$. The feature classifier is trained to correctly classify whether its inputs are from the same image or not, while the encoders are trained to fool the classifier. We employ the LSGAN framework~\cite{mao2017least} to update the networks because of its simplicity and fast convergence. We leave the application of other GAN variants such as WGANGP~\cite{gulrajani2017improved} in the future work. Specifically, we update the feature classifier with
\begin{eqnarray}
\mathcal{L}_{adv}(C_f) = (C_f(\Mat{z}_s^a, \Mat{z}_a^a)-1)^2 + (C_f(\Mat{z}_s^a, \Mat{z}_a^b))^2
\vspace{-3mm}
\end{eqnarray}
Conversely, we update encoders to output features that can 'fool' the classifier:
\begin{eqnarray}
\mathcal{L}_{adv}(E_s, E_a) = (C_f(\Mat{z}_s^a, \Mat{z}_a^a)-1)^2 + (C_f(\Mat{z}_s^a, \Mat{z}_a^a))^2
\vspace{-3mm}
\end{eqnarray}

\paragraph{Color Consistency loss} We add this regularization term to our framework to ensure the appearance statistics remain unchanged when transferred to other shapes. In particular, we enforce color consistency between the filtered patches\footnote{In our case, we refer $\Mat{I}\Mat{z}_{s, i}$ as the $i$-th image patch, as opposed to the conventional definition of image patch', which often refer to a cropped part.} of the generated image and those of the original image.
Let us define $x_n$ to be the $n$th pixel in the $i$th image patch $\Mat{I}\Mat{z}_{s, i}$, we can then calculate the mean $\mu$ and variance $\sigma$ of that sample by $\mu_i=\sum_\mathcal{N} x_n/N$ and $\sigma_i=\sum_\mathcal{N} (x_n-\mu_i)^2/N$ where $\mathcal{N}$ is the total number of pixels in the image.

We then impose color consistency by minimizing the squared difference of the color statistics:
\begin{equation}
\mathcal{L}_C = \frac{1}{m}\sum_i^m ((\mu_i^{mix}-\mu_i^a)^2 + (\sigma_i^{mix}-\sigma_i^a)^2)
\end{equation}
where $m$ is the number of shape masks.

\paragraph{Full Loss}
Putting above losses altogether, we attain the full loss as
\begin{equation}
\mathcal{L} = \lambda_1\mathcal{L}_R + \lambda_2\mathcal{L}_{adv}+ \lambda_3\mathcal{L}_C
\label{Eq:full_loss}
\end{equation}
where $\lambda_1, \lambda_2$ and $\lambda_3$ are hyper-parameters to control the weights of the three components. We sample mini-batches of randomly chosen image pairs $(\Mat{I}^a,\Mat{I}^b)$ to train our network.
\vspace{-2mm}
\section{Experiments}
In this section, we demonstrate the disentanglement ability of the proposed method which enable us to individually control each component of the generated image, namely the appearance and shape. We first evaluate the results on conditional person image generation in Section~\ref{sec.tansfer_results}. Then we explain and visualize the disentangled representations learned by our network. After that, we present both qualitative and quantitative comparisons with current state-of-the-art algorithms, showing that we achieve favourable results against other weakly or even fully-supervised methods. Finally, an ablation study is conducted. Two datasets are applied: DeepFashion~\cite{liuLQWTcvpr16DeepFashion} and Market1501~\cite{zheng2015scalable}.

\subsection{Datasets \& Implementation Details}
\textbf{DeepFashion}. We use the In-shop Clothes Retrieval Benchmark of DeepFashion which contains 52712 of in-shop clothes images. The dataset offers various kinds of clothes on many different persons. Following the previous protocol~\cite{esser2018variational}, we filter out invalid images with no person present, and then randomly select 32988 images for training and 8245 images for testing.

\textbf{Market1501}. The Market1501 dataset contains 322,668 images collected from 1501 persons. Each image is of spatial size 128x64. Joining the prior practice~\cite{esser2018variational}, we randomly utilize 9399 images for training and 3810 images for testing.

\textbf{Implementation Details}. The network is optimized using Adam optimizer \cite{kingma2014adam} with initial learning rate of 0.0001, $\beta_1=0.9$ and $\beta_2=0.999$. The learning rate is decreased by 5\% every 2500 iterations, also, learning rate for the shape encoder $E_s$ is further modified by a factor of 0.1. The batch size is set to 12 primarily due to hardware limitations. We update the generator($E_s$, $E_a$ and $D_I$) and the classifier ($C_f$) alternately on every step. The hyper-parameters in Eq~\ref{Eq:full_loss} are set to be: 
\begin{eqnarray}
\lambda_1=0.01,\ \lambda_2=1,\ \lambda_3=1
\vspace{-4mm}
\end{eqnarray}
We initialize the shape encoder network with pretrained weights of OpenPose net \cite{cao2017realtime} for stable training and fast convergence.

\subsection{Appearance \$ Shape Transfer}\label{sec.tansfer_results}
We show that the proposed network can independently control the appearance and shape of an image based only on two conditional images, each providing cues for one the attributes.
\begin{figure*}[t]
	\begin{center}
		\includegraphics[width=0.9\linewidth]{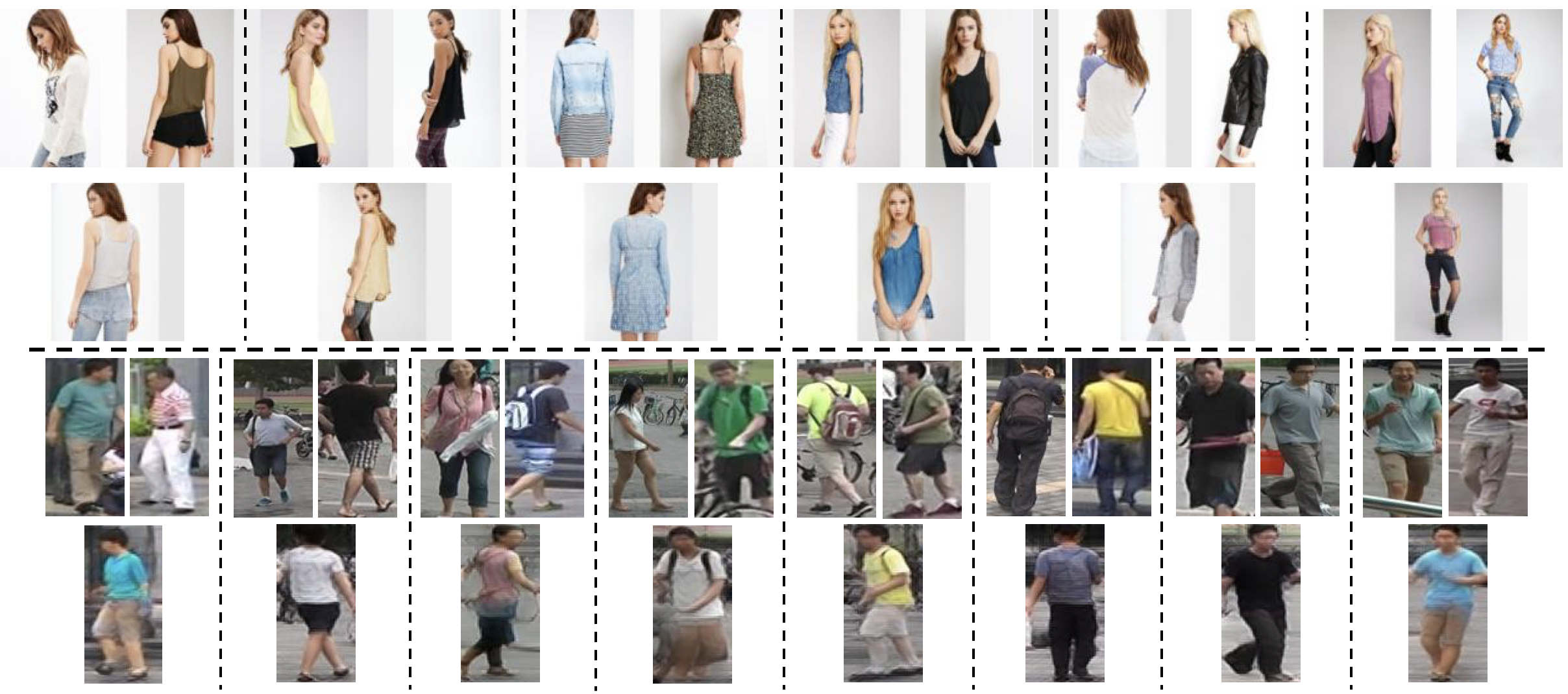}
	\end{center}
	\vspace{-3mm}
	\caption{Appearance and shape transfer results on DeepFashion (up) and market1501 (bottom). The input images are all randomly chosen. Images of upper/whole body, side/front/back views, are shown here to demonstrate the robustness of our method. In each block, the appearance is inferred from the upper-left image, while shape is inferred from the upper-right one.}
	\label{fig.deepfashion_market_result}
	\vspace{-3mm}
\end{figure*}

\begin{figure*}[t]
	\begin{center}
		\includegraphics[width=0.85\linewidth]{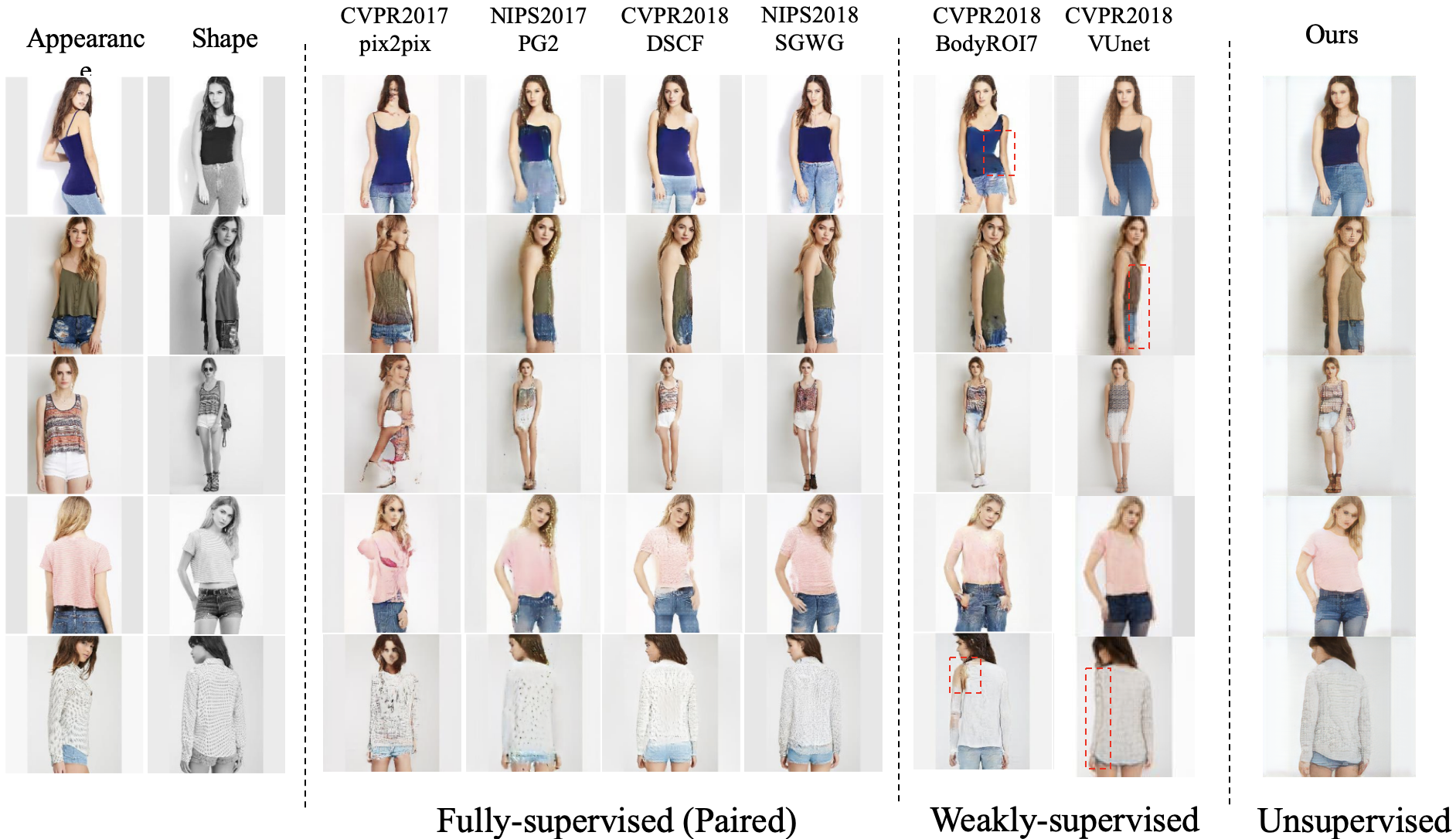}
	\end{center}
	\vspace{-3mm}
	\caption{Visual comparison with current state-of-the-arts on DeepFashion. We can see that our unsupervised method achieves better visual results compared to existing weakly-supervised methods. For fair comparison, our results are generated using gray-scale images as shape source. \textcolor{red}{Red boxes} points put areas of significant improvements of our methods.}
	\label{fig.compare_deepfashion}
	\vspace{-3mm}
\end{figure*}

\begin{figure}[!t]
	\begin{center}
		\includegraphics[width=1.0\linewidth]{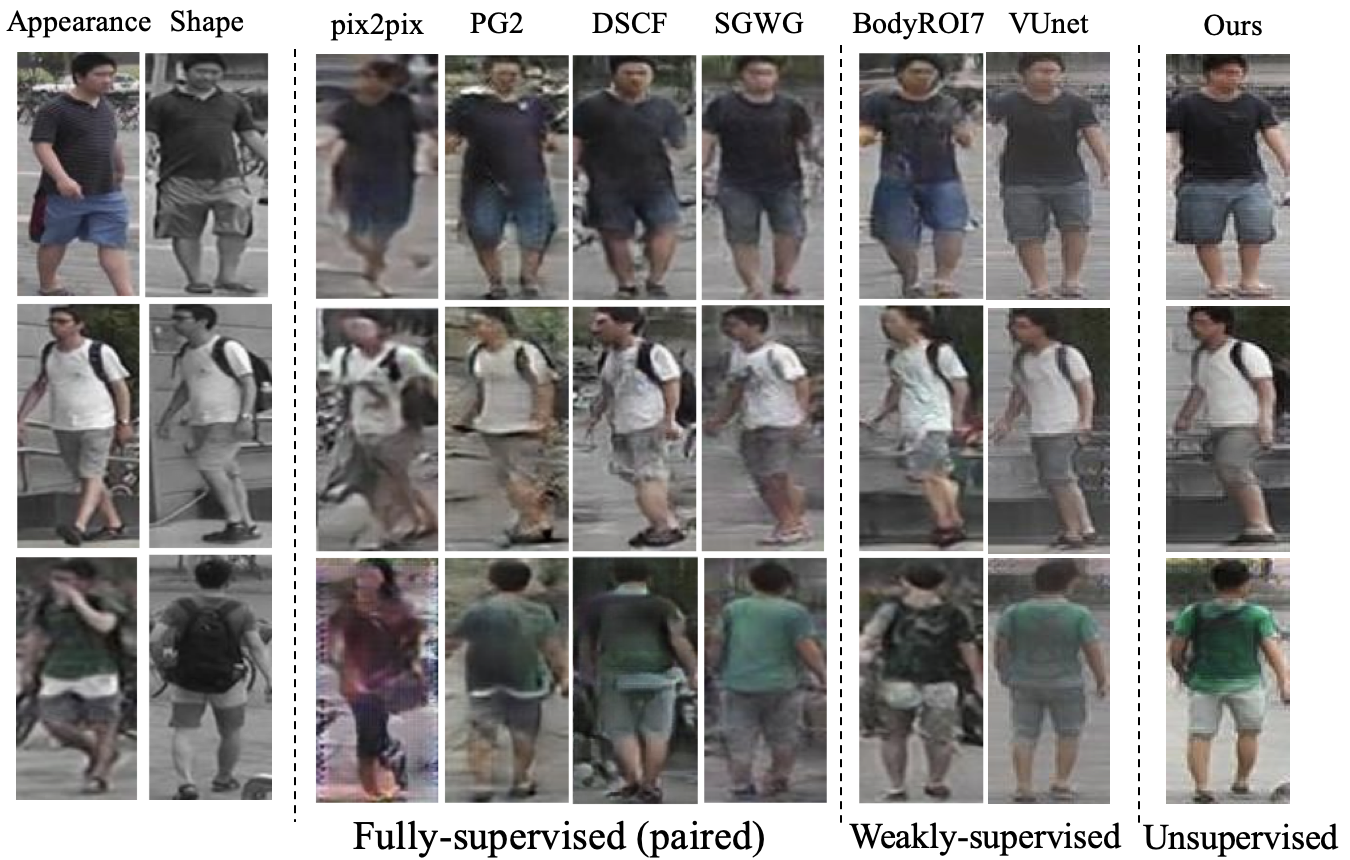}
	\end{center}
	\vspace{-3mm}
	\caption{Visual comparison with current state-of-the-arts on Market1501.}
	\label{fig.compare_market}
	\vspace{-3mm}
\end{figure}

Figure~\ref{fig.deepfashion_market_result} shows the image generation results on DeepFashion and market1501, where a mixture of upper/whole body images from side/front/back views are presented to show the robustness of our method. We can see the generated image preserves the conditional shape while being consistent with the appearance in color. Even some detailed shape characteristics are successfully preserved during the transfer process. For example, in the second block of Deepfashion images, it can be observed that the hair style in the generated image is as natural as in the original one. Also the details around the corner of the cloth are well preserved. 

\begin{figure}[t]
	\begin{center}
		\includegraphics[width=1.0\linewidth]{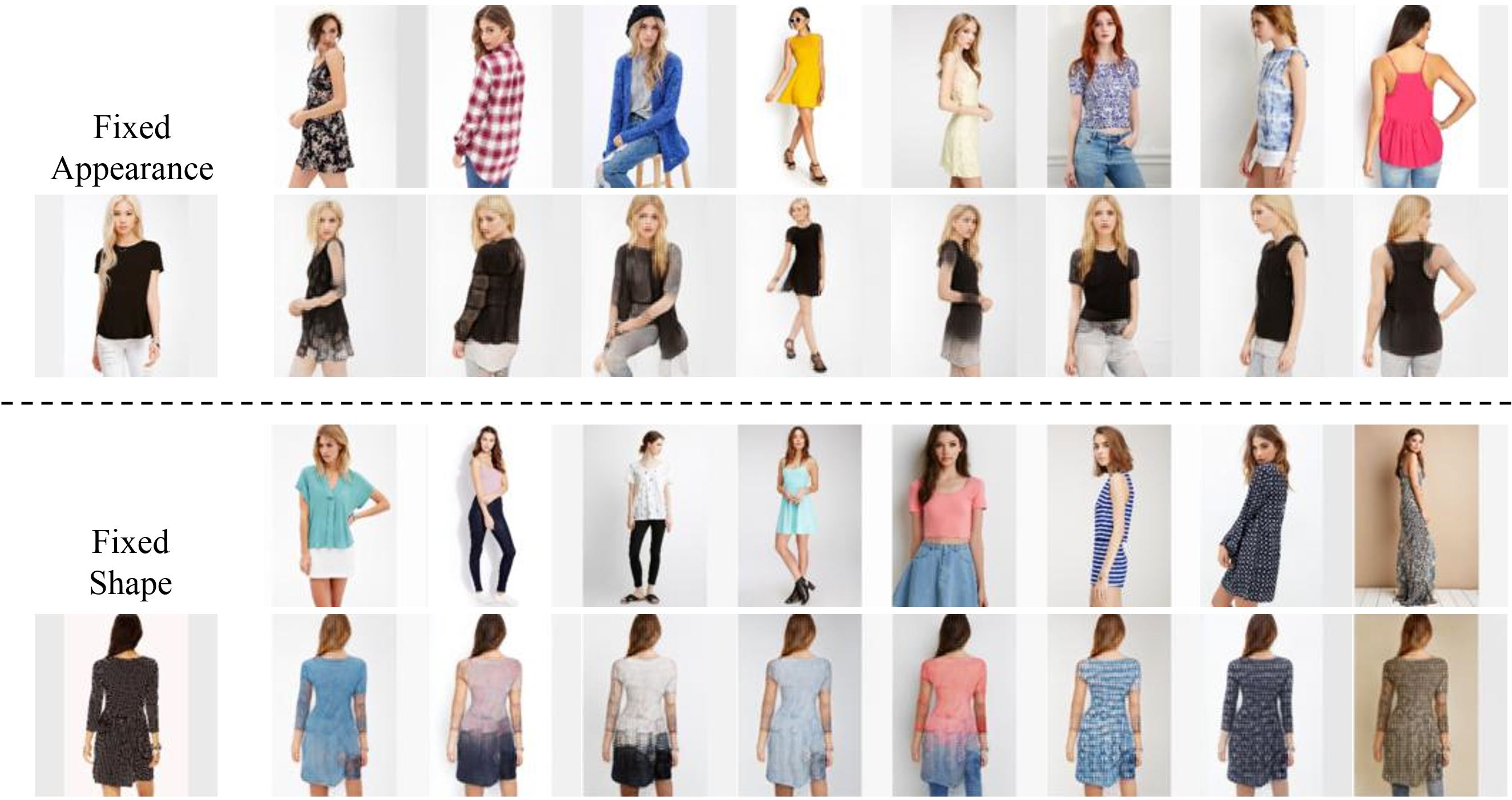}
	\end{center}
	\vspace{-3mm}
	\caption{Demonstration of model stability. Upper: same appearance paired with different shape; Lower: same shape with different appearance. We can see the desired image attributes are well-preserved and consistent in both cases.}
	\label{fig.stablity_deepfashion}
	\vspace{-3mm}
\end{figure}

To demonstrate the stability of our model, transfer results using the same appearance image with multiple shape images (and the reverse) are shown in Figure~\ref{fig.stablity_deepfashion}.
It is obvious that both the shape and appearance remain consistent and well-aligned with the conditional image across different generation results, proving good stability of the model.

\subsection{Understanding The Disentangled Representation}
\begin{figure}[t]
	\begin{center}
		\includegraphics[width=0.85\linewidth]{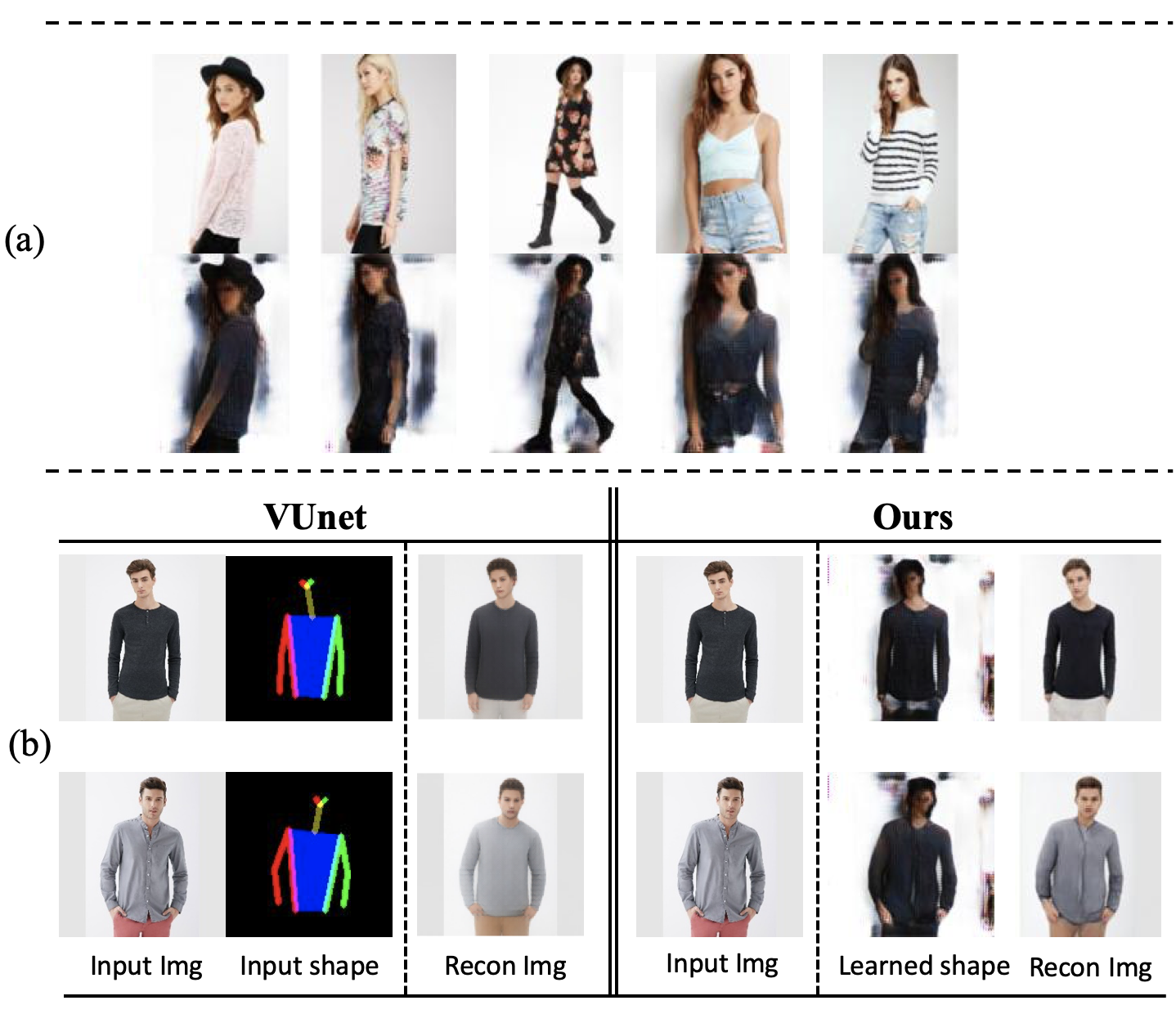}
	\end{center}
	\vspace{-3mm}
	\caption{Visualization of the learned shape masks and its effect. (a) visualize the learned shape masks by decoding $\Mat{z}_{s,i}$ with all-zero appearance features. (b) demonstrates the advantages of our learned masks over fixed stickman inputs employed in~\cite{esser2018variational}. Notice how results from VUnet are unnatural because the input stickman enforce too much human bias onto their network.}
	\label{fig.mask_details}
	\vspace{-3mm}
\end{figure}

As explained above, we use pre-trained OpenPose net as our shape prior, and adaptively refine the generated shape masks. The shape masks not only guide the appearance encoder the encoder local patterns, but also serve as shape conditions for the image decoder. Figure~\ref{fig.mask_details} shows the details of shape mask refinement, as well as illustrate the positive effect it gives to image generation.

We can see from Figure~\ref{fig.mask_details} (a) that shape masks changed dramatically after refinement. Although some background noise got encoded into the refined masks, they conform around the input shape much more tightly than the original ones, \emph{e.g.} we can clearly see that the bottom left mask focuses on the lower legs and shoes while the original one only roughly indicates their location. Such mask refinement offer noticeable improvements in the quality of the generated images, as shown in the two reconstructed images on the right side. Figure~\ref{fig.mask_details} (b) shows better visualization of the learned shape representation. In our framework, the shape encoder $E_s$ generates a set of 14 masks, each focusing on a different body part. To better visualize the combined effect of all the masks, we decode images by setting the appearance features to all zeros, thus obtain clean shape visualizations of shape representation, which is shown in bottom row of (b). It is obvious that our network nicely preserve the input shape, including even small details such as hats (first column) and shoes (third column). Figure~\ref{fig.mask_details} (c) demonstrates the advantages offered by learning shape representations over using explicit shape cues. Comparing the reconstruction results between ours and \cite{esser2018variational}, our images exhibits much more details and are hardly distinguishable to the original ones. Because our learned shape masks conforms to the input human contour, details like the gaps between arms and torso are well preserved.

\subsection{Compare with Current State-Of-The-Art}
To the best of our knowledge, our proposed method is the first unsupervised end-to-end framework that achieves automatic disentanglement and image generation without heavy pre/post-processing. We compare with three methods that are closest to ours: BodyRoI~\cite{ma2018disentangled}, UPIS~\cite{pumarola2018unsupervised} and V-Unet~\cite{esser2018variational}, where they also achieve unpaired image translation, but weak supervision (joint coordinates) are provided. To put the performance of our method into context, we also compare with supervised (paired) image translation methods, specifically PG2~\cite{ma2017pose}, DeformableGan~\cite{siarohin2018deformable} and SGWG~\cite{dong2018soft}. As a general baseline, we also compare with pix2pix~\cite{isola2017image}. Experimental results shows that we compare favourably against all current state-of-the-arts, even supervised ones.

Visual comparisons are shown in Figure~\ref{fig.compare_deepfashion} and Figure~\ref{fig.compare_market} for DeepFashion and Market1501 respectively. We can see that we generate the most visual pleasing and realistic results with well-preserved details. We also conduct quantitative comparisons evaluated with SSIM score. Results on both datasets are reported in Table~\ref{table.quantitative_comparison}, suggesting superior performance of our methods.
\begin{table}[ht]
	\begin{center}
		\begin{tabular}{ l  c  c  c  c } 
	     \hline
         & \multicolumn{2}{c}{DeepFashion} & \multicolumn{2}{c}{Market1501}\\
         \hline
         \textbf{Methods} & SSIM & IS & SSIM & IS \\
         \hline
         pix2pix\cite{isola2017image} & 0.692 & 3.249 & 0.183 & 2.678 \\
         PG2\cite{ma2017pose} & 0.762 & 3.090 & 0.253 & 3.460 \\
         DSCF\cite{siarohin2018deformable} & 0.761 & \textbf{3.351} & 0.290 & 3.185 \\
         SGWG\cite{dong2018soft} & 0.793 & 3.314 & 0.356 & 3.409 \\ 
         \hline
         \hline
         UPIS\cite{pumarola2018unsupervised} & 0.747 & 2.97 & - & - \\
         BodyROI7\cite{ma2018disentangled} & 0.614 & 3.228 & 0.099 & \textbf{3.483} \\
         VUnet\cite{esser2018variational} & 0.786 & 3.087 & 0.353 & 3.214 \\
         \hline
         \hline
         Ours & \textbf{0.844} & 3.196 & \textbf{0.626} & 3.281 \\
         \hline
		\end{tabular}
	\end{center}
	\vspace{-1mm}
    \caption{Quantitative Comparison using SSIM and IS score. Following \cite{esser2018variational} and \cite{dong2018soft}, the SSIM score is calculated using reconstructed images in the test set. The first block are supervised methods using paired data; middle block are weakly-supervised methods requiring explicit shape inputs. Our unsupervised method achieve comparable scores against supervised or weakly-supervised methods}
    \label{table.quantitative_comparison}
    \vspace{-1mm}
\end{table}


\subsection{Ablation Study}
\begin{figure}[ht]
	\begin{center}
		\includegraphics[width=0.8\linewidth]{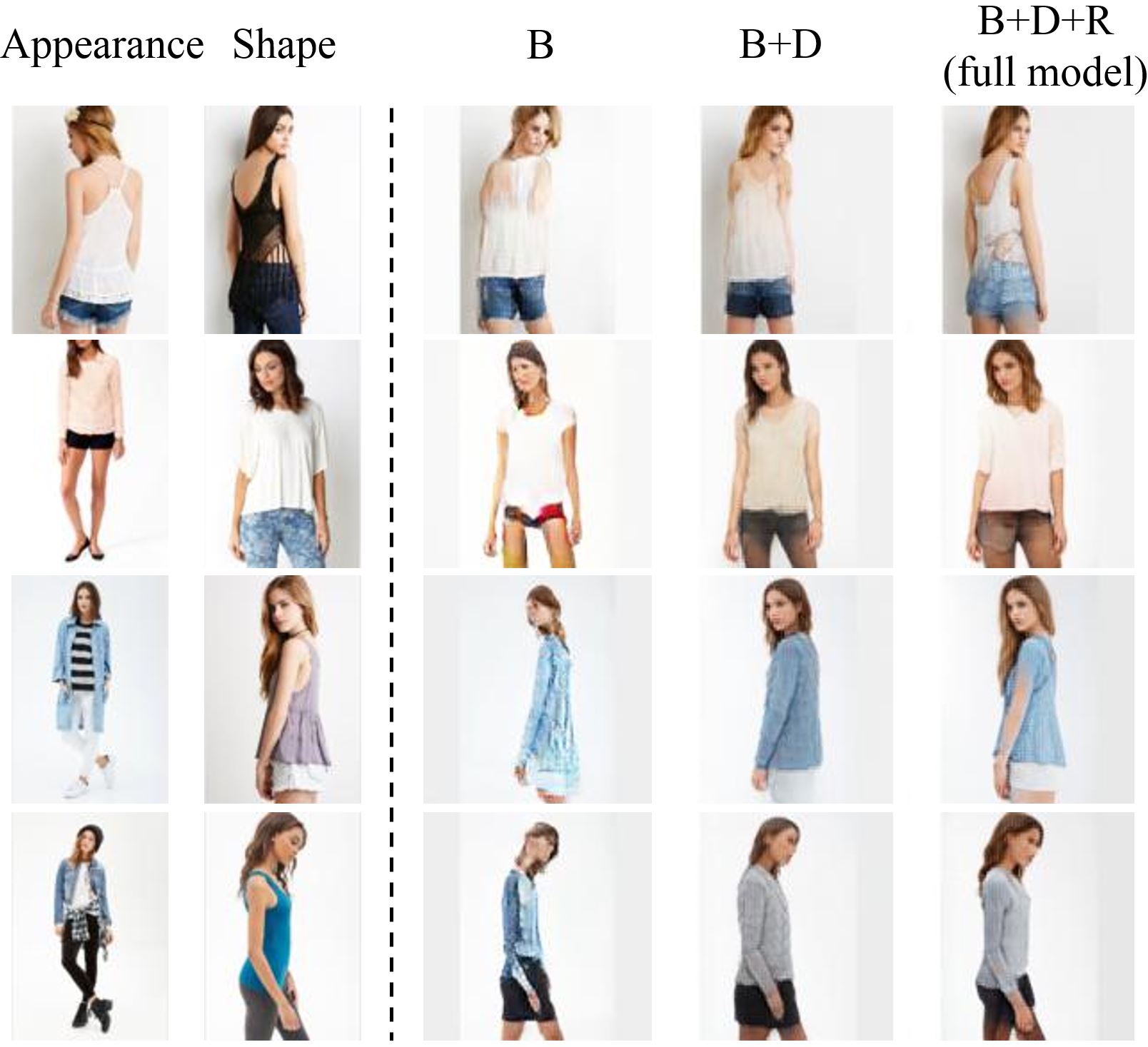}
	\end{center}
	\vspace{-3mm}
	\caption{Ablation study results comparing the performance of base model (B), base model with feature disentanglement (B+D), and our full model using both disentanglement and shape mask refinement (B+D+R).}
	\label{fig.ablation_study}
	\vspace{-3mm}
\end{figure}

We argue that the two most important contributing factors to our superior performance is \emph{a learned more plausible shape representation} and \emph{clean disentangled features} encouraged by feature adversarial and color consistency loss. We conduct an ablation study to verify their claimed effects. Figure~\ref{fig.ablation_study} shows results using three model variations.
In the above figure, base model is the bare-bone network using fixed shape encoder trained with perceptual reconstruction loss only. The results indicates that, without feature disentanglement, the appearance contain much information about its original shape, yielding significant visual artifacts and incoherent human images. After adding the feature adversarial loss and color consistency loss to encourage disentanglement, results are noticeably better with different appearance patterns in the correct places. The performance are further improved by employing adaptive shape refinement in our full model. 
Refined shape offer much more details, such as body curvature, hair styles etc.

\vspace{-1mm}
\section{Conclusion}
In conclusion, We have developed an unsupervised framework that learns to extract latent representations of appearance and shape, as well as to generate novel images from unpaired data. Compared to existing methods using fixed shape cues, our learned shape representation encodes finer details, yielding more realistic images. Experiments results reports superior performance quantitatively and visually against current state-of-the-arts, including even some supervised ones.

{\small
\bibliographystyle{ieee}
\bibliography{egbib}
}

\end{document}